\documentclass[10pt,twocolumn,letterpaper]{article}

\usepackage{cvpr}
\usepackage{times}
\usepackage{epsfig}
\usepackage{graphicx}
\usepackage{amsmath}
\usepackage{amssymb}
\usepackage{filecontents,url}
\usepackage{subfig}

\cvprfinalcopy 

\def\cvprPaperID{5} 

\begin{document}

\title{A Coarse-to-fine Deep Convolutional Neural Network Framework for Frame Duplication Detection and Localization in Forged Videos}

\author{Chengjiang Long ~~~~~~ Arslan Basharat ~~~~~~ Anthony Hoogs\\
Kitware Inc.\\
28 Corporate Drive, Clifton Park, NY 12065\\
{\tt\small \{chengjiang.long, arslan.basharat, anthony.hoogs\}@kitware.com}
}

\maketitle

\begin{abstract}
Videos can be manipulated by duplicating a sequence of consecutive frames with the goal of concealing or imitating a specific content in the same video. In this paper, we propose a novel coarse-to-fine framework based on deep Convolutional Neural Networks to automatically detect and localize such frame duplication. First an I3D network finds coarse-level matches between candidate duplicated frame sequences and the corresponding selected original frame sequences. Then a Siamese network based on ResNet architecture identifies fine-level correspondences between an individual duplicated frame and the corresponding selected frame. We also propose a robust statistical approach to compute a video-level score indicating likelihood of manipulation or forgery. Additionally, for providing manipulation localization information we develop an inconsistency detector based on the I3D network to distinguish the duplicated frames from the selected original frames. Quantified evaluation on two challenging video forgery datasets clearly demonstrates that this approach performs significantly better than four recent state-of-the-art methods.
\end{abstract}

\section{Introduction}
\begin{figure}
\begin{center}
\includegraphics[width=0.45\textwidth]{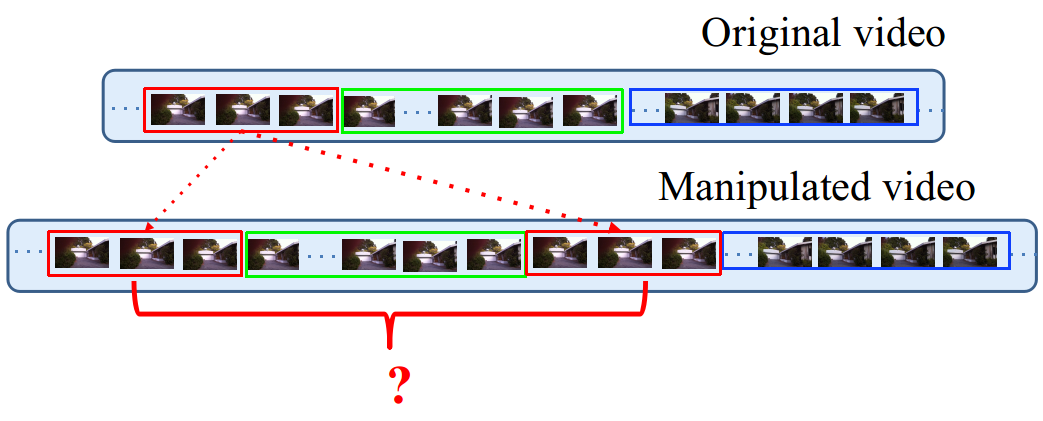}
\end{center}
\vspace{-0.6cm}
   \caption{An illustration of frame duplication manipulation in a video. Assume an original video has three sets of frames indicated here by red, green and blue rectangles. A manipulated video can be generated by inserting a second copy of the red set in the middle of the green and the blue sets. Our goal is to detect both instances of the red set as duplicated and also determine that the second instance is the one that's forged.}
\vspace{-0.5cm}   
\label{fig:problem}
\end{figure}

An increasingly large volume of digital video content is becoming available in our daily lives through the internet due to rapid growth of increasingly sophisticated, mobile and low-cost video recorders. These videos are often edited and altered for various purposes using image and video editing tools that have become more readily available. Manipulations or forgeries can be done for nefarious purposes to either hide or duplicate an event or content in the original video. Frame duplication refers to a video manipulation where a copy of a sequence of frames inserted into the same video either replacing previous frames or as additional frames. Figure~\ref{fig:problem} provides an example of frame duplication where in the manipulated video the red frame sequence from the original video is inserted between the green and the blue frame sequences. As a real-world example, frame duplication forgery could be done to hide an individual leaving a building in a surveillance video. If such a manipulated video was part of a criminal investigation, without effective forensics tools the investigators could be misled. 

It is very important to develop robust video forensic techniques, like the one proposed here, to catch videos with increasing sophisticated forgeries. Video forensics techniques~\cite{milani2012overview,wang2007exposing} aim to extract and exploit features from videos that can distinguish forgeries from original, authentic videos. Like other areas in information security the sophistication of attacks and forgeries continue to increase for images and videos, requiring a continued improvement in the forensic techniques. Robust detection and localization of duplicated parts of a video can be a very useful forensic tool for those tasked with authenticating large volumes of video content.

In recent years, multiple digital video forgery detection approaches have been employed to solve this challenging problem. Wang and Farid~\cite{wang2007exposing} proposed a frame duplication detection algorithm which takes the correlation coefficient as a measure of similarity. However, such an algorithm requires heavy computational load due to a large amount of correlation calculations. Lin {\em et al.}~\cite{lin2012detection} proposed to use histogram difference (HD) instead of correlation coefficients as the detection features.  The drawback is that the HD features do not show strong robustness against common video operations or attacks.  Hu {\em et al.}~\cite{hu2012improved} propose to detect duplicated frames using video sub-sequence fingerprints extracted from the DCT coefficients. Yang {\em et al.}~\cite{yang2016using} propose an effective similarity-analysis-based method
that is implemented in two stages, where the features are obtained via SVD. Ulutas {\em et al.} propose to use a BoW model~\cite{ulutas2018frame} and binary features~\cite{ulutas2017frame} for frame duplication detection. Although deep learning solutions, especially those based on convolution neural networks, have demonstrated promising performance in solving many challenging vision problems such as large-scale image recognition~\cite{He_2016_CVPR, Stock_2018_ECCV}, object detection~\cite{NIPS2015_5638, Chen_2018_CVPR, Tang_2018_ECCV} and visual captioning~\cite{VenugopalanXDRMS14, Aneja_2018_CVPR, Yu_2018_CVPR}, no deep learning solutions have been developed for this specific task so far, which motivates us to fill this gap.

\begin{figure*}
\begin{center}
\includegraphics[width=0.99\textwidth]{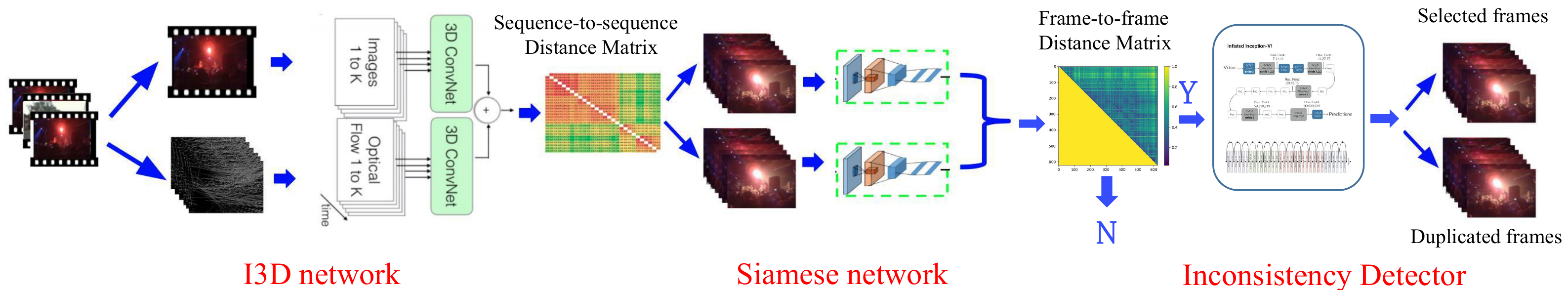}
\end{center}
\vspace{-0.6cm}
 \caption{The proposed C2F-DCNN framework for frame duplication detection and localization. Given a testing video, we first run the I3D network~\cite{carreira2017quo} to extract deep spatial-temporal features and build the coarse sequence-to-sequence distance to determine the possible frame sequences that are likely to have frame duplication. For the likely duplicated sequences, a ResNet-based Siamese network further confirms a frame duplication at frame-level. For the videos with duplication detected, temporal localization is determined with an I3D-based inconsistency detector to distinguish the duplicated frames from the selected frames.}
\label{fig:framework}
\vspace{-0.3cm}
\end{figure*}

In this paper, we propose a novel coarse-to-fine deep learning framework, denoted as C2F-DCNN, for frame duplication detection and localization in forged videos. As illustrated in Figure~\ref{fig:framework}, we first utilize an I3D network~\cite{carreira2017quo} to obtain the candidate duplicate sequences at a coarse level; this helps narrow the search faster through longer videos. Next, at a finer-level, we apply a Siamese network composed of two ResNet networks~\cite{He_2016_CVPR} to further confirm duplication at the frame level to obtain accurate corresponding pairs of duplicated and selected original frames. Finally, the duplicated frame-range can be distinguished from the corresponding selected original frame-range by our inconsistency detector that is designed as a I3D network with 16-frames as a input video clip.


Unlike other methods, we consider the consistency between two consecutive frames from a 16-frame video clip in which these two consecutive frames are in center, {\em i.e.}, 8-th and 9-th frames. This is aimed at capturing the temporal-context for matching a range of frames for duplication. Inspired by Long {\em et al.}~\cite{Long:CVPRW2017}, 
we design an inconsistency detector based on the I3D network to cover three categories, {\em i.e.}, ``none", ``frame drop", and ``shot break", which represent that between 8-th and 9-th frames there are no manipulations, there are frames removal within one shot, and there exist two shots transit in the 16-frame video clips, respectively. 
Therefore, we are able to use output scores from the learned I3D network to formulate a confidence score of inconsistency between any two consecutive frames to distinguish the duplicated frame-range from the selected original frame-range, even in videos with multiple shots.

We also propose a heuristic strategy to produce a video-level frame duplication likelihood score. This is built upon the measures like number of possible frames duplicated, minimum distance between duplicated frames and selected frames, and the temporal gap between the duplicated frames and the selected original frames.

To summarize, the contributions of this paper are as follows: 
\begin{itemize}
\item We propose a novel coarse-to-fine deep learning framework for frame duplication detection and localization in forged videos. This framework features fine tuned I3D networks and the ResNet Siamese network, providing a robust yet efficient approach to process large volumes of video data.
\item We have designed an inconsistency detector based on a fine-tuned I3D network that covers three categories to distinguish duplicated frame-range from the selected original frame-range.
\item We propose a heuristic formulation for video-level detection score, which leads to significant improvement in detection benchmark performance.
\item We evaluate performance on two video forgery datasets and the experimental results strongly demonstrate the effectiveness of the proposed method.
\end{itemize}

\section{Related Work}
The research related to frame duplication can be broadly divided into {\em inter-frame forgery}, {\em copy-move forgery} and {\em Convolutional Neural Networks}.


{\bf Inter-frame forgery} refers to frame deletion and frame duplication. 
For features used for inter-frame forgery, either spatially or temporally, keypoints are extracted from nearby patches recognized over distinctive scales. 
Keypoint-based methodologies can be further subdivided into direction based~\cite{DouzeGJMS08, le2010national}, keyframe-based coordinating~\cite{law2006robust} and visual words based~\cite{sowmya2015survey}. 
In particular, keyframe-based feature has been indicated to display incredible execution for close video picture/feature identification~\cite{law2006robust}. 

In addition to keypoint-based features Wu {\em et al.}~\cite{wu2014exposing} proposes a velocity field consistency based approach to detect inter-frame forgery. 
This method is able to distinguish the forgery types, identify the tampered video and locate the manipulated positions in forged videos as well. 
Wang {\em et al.}~\cite{wang2014videoJCC} propose to make full use of the consistency of the correlation coefficients of gray values to classify original videos and inter-frame forgeries. 
They also propose an optical flow method~\cite{wang2014videoST} based on the assumption that the optical flows are consistent in an original video, while in forgeries the consistency will be destroyed. The optical flow is extracted as distinguishing feature to identify inter-frame forgeries through a Support Vector Machine (SVM) classifier to recognize frame insertion and frame deletion forgeries.

Recently, Huang {\em et al.}~\cite{huang2018multi} proposed a fusion of audio forensics detection methods for video inter-frame forgery. Zhao {\em et al.}~\cite{zhao2018inter} developed a similarity analysis based method to detect inter-frame forgery in a video shot. In this method, the HSV color histogram is calculated to detect and locate tampered frames in the shot, and then the SURF feature extraction and FLANN (Fast Library for Approximate Nearest Neighbors) matching are used for further confirmation.


{\bf Copy-move forgery} is created by copying and pasting content within the same frame, and potentially post-processing it~\cite{christlein2012evaluation, d2019patchmatch}. 
Wang {\em et al.}~\cite{wang2009fast} propose a dimensionality reduction approach through PCA (Principal Component Analysis) on the different pieces. 
The drawback is that for dark scale pictures furthermore forms each shading direct in shading pictures and PCA is for recognition the fakes. 
Mohamadian {\em et al.}~\cite{mohamadian2013detection} develop a Singular Value Decomposition (SVD) based method in which the image is isolated into numerous little covering squares and after that SVD is requested to remove the copied frames. Its shortcoming is that the method is not for shading pictures.

Recently, Yang {\em et al.}~\cite{yang2018copy} proposed a copy-move forgery detection based on a modified SIFT-based detector. Wang {\em et al.}~\cite{wang2018robust} presented a novel block-based robust copy-move forgery detection approach using invariant quaternion exponent moments. D`Amiano {\em et al.}~\cite{d2019patchmatch} proposed a dense-field method with a video-oriented version of PatchMatch for the detection and localization of copy-move video forgeries.


{\bf Convolutional Neural Networks} (CNNs) have been demonstrated to learn rich, robust and powerful features for large-scale video classification~\cite{karpathy2014large}. Various 3D CNN architectures~\cite{tran2015learning, carreira2017quo, Hara_2018_CVPR, Xie_2018_ECCV} have been proposed to explore spatio-temporal contextual relations between consecutive frames for representation learning. Unlike the existing methods for inter-frame forgery and copy-move forgery which mainly use hand-crafted features or Bag-of-Words, we take advantage of Convolutional Neural Networks to extract spatial and temporal features for frame duplication detection and localization.

\section{Proposed Approach}
As shown in the Figure~\ref{fig:framework}, given a probe video, our proposed C2F-DCNN framework is designed to detect and localize frame duplication manipulation. An I3D network is used to produce sequence-to-sequence matrix and determine the candidate frame sequences at the coarse-search stage. A Siamese network is then applied for a fine-level search to verify whether frame duplications exist. After this an inconsistency detector is applied to further distinguish duplicated frames from selected frames. All of these steps are described below in detail.

\subsection{Coarse-level Search for Duplicated Frame Sequences}\label{sec:coarse}
In order to efficiently narrow the search space, we start by finding possible duplicate sets of frames throughout the video using a robust CNN representation. We split a video into overlapping frame sequences, where each sequence has 64 frames and the number of overlapped frames is 16. We choose I3D Network~\cite{carreira2017quo}, instead of using C3D network~\cite{tran2015learning} due to these reasons: (1) it inflates 2D ConvNets into 3D and makes filters from typically $N\times N$ square to N$\times$N$\times$N cubic; (2) it bootstraps 3D filters from 2D filters to bootstrap parameters from the pre-trained ImageNet models, and (3) it paces receptive field growth in space, time and network depth.

In this paper, we apply the pre-trained off-the-shell I3D network to extract the 1024-dimensional feature vector for $k=64$ frame sequences since the input for the standard I3D network is 64 rgb-data and 64 flow-data. 
We observed that a lot of time was being spent on the pre-processing, so improved the testing runtime. First $k$ rgb-data and $k$ flow-data items are computed, then for the next frame sequence, we can copy $(k-1)$ rgb-data and $(k-1)$ flow-data from the previous video clip, and only calculate the last rgb-data and flow-data. This significantly improved the testing efficiency.

Based on the sequence features, we calculate the sequence-to-sequence distance matrix over the whole video using L2 distance. If the distance is smaller than the threshold $T_1$, then this indicates that these two frame sequences are likely duplicated and we take them as two candidate frame sequences for further confirmation during the next fine-level search. 

\subsection{Fine-level Search for Duplicated Frames}
For the candidate frame sequences, detected by the previous stage described in Section~\ref{sec:coarse}, we evaluate the distance between all pairs of frames across the two sequences, {\em i.e.}, a duplicated frame and the corresponding selected original frame. For this purpose we propose a Siamese neural network architecture, which learns to differentiate between two frames in the provided pair. It consists of two identical networks by sharing exactly the same parameters, each taking one of the two input frames. A contrastive loss function is applied to the last layers to calculate the distance between the pair. In principle, we can choose any neural network to extract feature for each frame. 

\begin{figure}
\begin{center}
\includegraphics[width=0.35\textwidth]{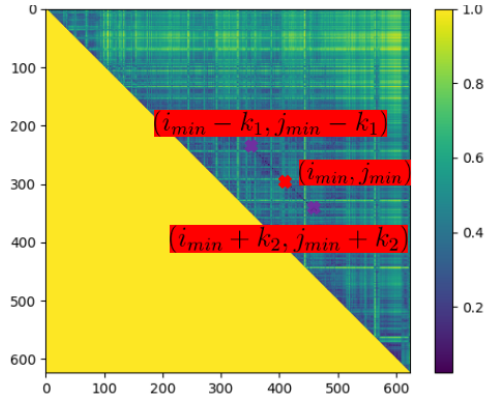}
\end{center}
\vspace{-0.75cm}
\caption{A sample distance matrix based on the frame-to-frame distances computed by the Siamese network between a pair of frame sequences. The symbols shown on the line segment with low distance are used to compute the video-level confidence score for frame duplication detection.}
\label{fig:videoscore}
\vspace{-0.3cm}
\end{figure}

In this paper, we choose the ResNet network~\cite{He_2016_CVPR} with 152 layers given its demonstrated robustness. We connect two ResNets in the Siamese architecture with a contrastive loss function and each loss value associated with the distance between a pair of frames is formulated into the frame-to-frame distance matrix, in which the distance is normalized to the range [0, 1]. A distance smaller than the threshold $T_2$ indicates that these two frames are likely duplicated. For videos that have multiple consecutive frames duplicated we expect to see a line with low values parallel to the diagonal in the visualization of the distance matrix, as plotted in Figure~\ref{fig:videoscore}.

It is worth mentioning that we provide both frame-level and video-level score to evaluate the likelihood of frame duplication. For the frame-level score, we can use the value in the frame-to-frame distance directly. For the video-level score, we propose a heuristic strategy to formulate the confidence value. We first find the minimal value of distance $d_{min} = d(i_{min}, j_{min})$ where $i_{min}, j_{min} = \operatorname*{argmin}\limits_{0 \leq i < j \leq n}d(i,j)$ in the frame-to-frame distance matrix. Then a search in performed in two directions to find the number of consecutive duplicated frames:
\begin{equation}
k_1 = \operatorname*{argmax}_{k: k\le i_{min}}
|d(i_{min} - k, j_{min} -k) - d_{min} | \leq \epsilon \\ 
\end{equation}
and
\begin{equation}
k_2 = \operatorname*{argmax}_{k:k\le n - j_{min}}
|d(i_{min} + k, j_{min} + k) - d_{min} | \leq \epsilon \\ 
\end{equation}
where $\epsilon = 0.01$ and the length of the interval with duplicated frames can be defined as:
\begin{equation}
l = k_1 + k_2 + 1.
\end{equation}

\noindent Finally, we can formulate the video-level confidence score as follows:
\begin{equation}
F_{video} = -\frac{d_{min}}{l \times (j_{min} - i_{min})} \label{eqn:vconfscore}
\end{equation}
The intuition here is that a more likely frame duplication is indicated by a smaller value of $d_{min}$, a longer interval of duplicated frames, and a larger temporal gap between the selected original frames and the duplicated frames.

\subsection{Inconsistency Detector for Duplication Localization}

\begin{figure}
\begin{center}
\includegraphics[width=0.35\textwidth]{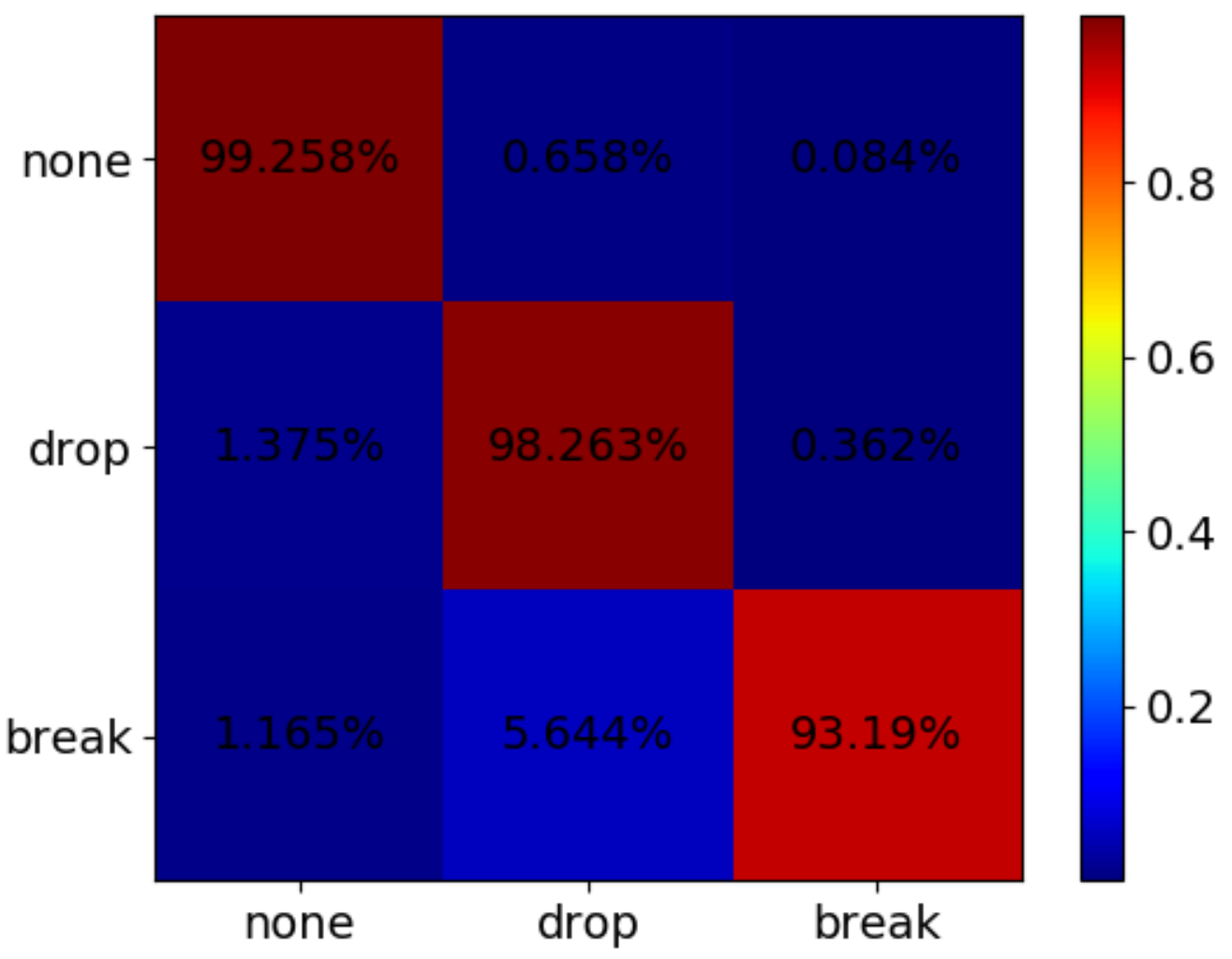}
\end{center}
\vspace{-0.5cm}
\caption{The confusion matrix for three classes of temporal inconsistency within a video, used with the I3D-based inconsistency. We expect high likelihood of ``drop'' class at the two ends of the duplicated frame sequence and high ``none'' likelihood at the ends of the selected original frame sequence.}
\label{fig:i3dscores}
\vspace{-0.2cm}
\end{figure}

We observe that the duplicated frames inserted into the source video usually yield artifacts due to temporal inconsistency at both the beginning frames and the end frames in a manipulated video. To automatically distinguish the duplicated frames from selected frames, we make use of both spatial and temporal information by training an inconsistency detector to locate this temporal discrepancy. For this purpose, we build upon the work by Long {\em et al.}~\cite{Long:CVPRW2017} who proposed a C3D-based network for frame-drop detection and only works for single shot videos. 
Instead of using only one RGB stream data as input, we replace the C3D network with an I3D network to also incorporate the optical flow data stream. It's also worth mentioning that unlike the I3D network used in Section~\ref{sec:coarse}, input to the I3D network here is a 16-frame temporal interval, every frame in a sliding window, with RGB and optical flow data.
The temporal classification provides insight about the temporal consistency between the 8-th and the 9-th frame within the 16-frame interval.
In order to handle multiple shots in a video with hard cuts, we extend the binary classifier to three classes: ``none" - no temporal inconsistency indicating manipulation; ``frame drop" - there are frames removed within one shot video; and ``shot break" or ``break'' - there is a temporal boundary or transition between two video shots. 
Note that the training data with shot-break videos are obtained from TRECVID 2007 dataset~\cite{KawaiSY07}, and we only use the hard-cut shot-breaks since soft-cut changes gradually and has strong consistency between any two consecutive frames. The confusion matrix in Figure~\ref{fig:i3dscores} illustrates high effectiveness of the proposed I3D network based inconsistency detector.

\begin{figure}
\begin{center}
\includegraphics[width=0.475\textwidth, height=0.275\textwidth]{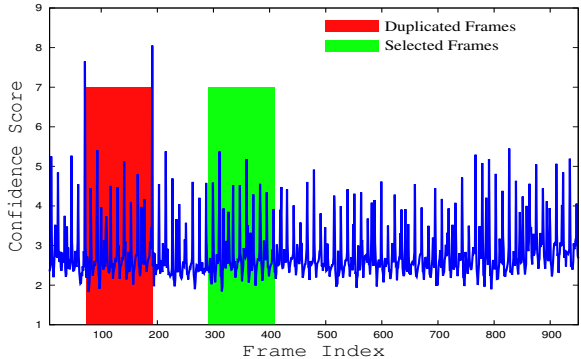}
\end{center}
\vspace{-0.5cm}
\caption{Illustration of distinguishing duplicated frames from the selected frames. The index ranges for the red frame sequence and the green sequence are [72, 191] and [290, 409], respectively. $s_1$ and $s_2$ are the corresponding inconsistency scores for red sequence and green sequence, respectively. Obviously, $s_1 > s_2$, which indicates that the red sequence is duplicated frames as expected.}
\label{fig:localization}
\vspace{-0.3cm}
\end{figure}

Based on the output scores for the three categories from the I3D network, {\em i.e.}, $S_{I3D}^{none}(i)$, $S_{I3D}^{drop}(i)$, and $S_{I3D}^{break}(i)$,  
we formulate the confidence score of inconsistency as the following function
\begin{equation}
\label{eq:i3d}
S(i) = S_{I3D}^{drop}(i) + S_{I3D}^{break}(i) - \lambda S_{I3D}^{none}(i),
\end{equation}
where $\lambda$ is the weight parameter, and for the results presented here we use $\lambda = 0.1$.
We assume the selected original frames have a higher temporal consistency with frames before and after such frames than the duplicated frames because the insertion of duplicated frames usually causes a sharp inconsistency at the beginning and the end of the duplicated interval, as illustrated in Figure~\ref{fig:localization}. Given a pair of frame sequences that are potentially duplicated, $[i, i+l]$ and $[j, j+l]$, we compare two scores,
\begin{equation}
s_1 = \sum\limits_{k=-wind}^{wind} S(i-1+k) + S(i+l+k)
\end{equation}
and
\begin{equation}
s_2 = \sum\limits_{k=-wind}^{wind} S(j-1+k) + S(j+l+k),
\end{equation}
where $wind$ is the window size we check the inconsistency at both the beginning and the end of the sequence. In this paper, we set $wind=3$ to avoid the failure cases where a few start or end frames were detected incorrectly. If $s_1 > s_2$, then the duplicated frame segment is $[i, i+l]$. Otherwise, the duplicated frame segmentation is $[j, j+l]$. As shown in Figure~\ref{fig:localization}, our modified I3D network is able to measure the consistency between consecutive frames. 

\section{Experimental Results}
We evaluate our proposed C2F-DCNN method on a self-collected video dataset and the Media Forensics Challenge 2018 (MFC18)\footnote{URL: https://www.nist.gov/itl/iad/mig/media-forensics-challenge-2018.} dataset~\cite{guan2019mfc}.

Our self-collected video dataset is obtained through taking frame duplication manipulation on the 12 
raw static camera videos from VIRAT dataset~\cite{oh2011large} and 17 
dynamic iPhone 4 videos. 
In order to generate test videos with frame duplication We randomly select frame sequences with the duration 0.5s, 1s, 2s, 5s and 10s, and then re-insert them into the same source videos. We use the X264 video codec and a frame rate of 30 fps to generate these manipulated videos. Note that we avoid any temporal overlap between the selected original frames and the duplicated frames in all generated video. Since we have the frame-level ground truth, we can use it for frame-level performance evaluation.

\begin{figure*}
\vspace{-0.10in}
\begin{center}
\includegraphics[width=0.90\textwidth,height=0.45\textwidth]{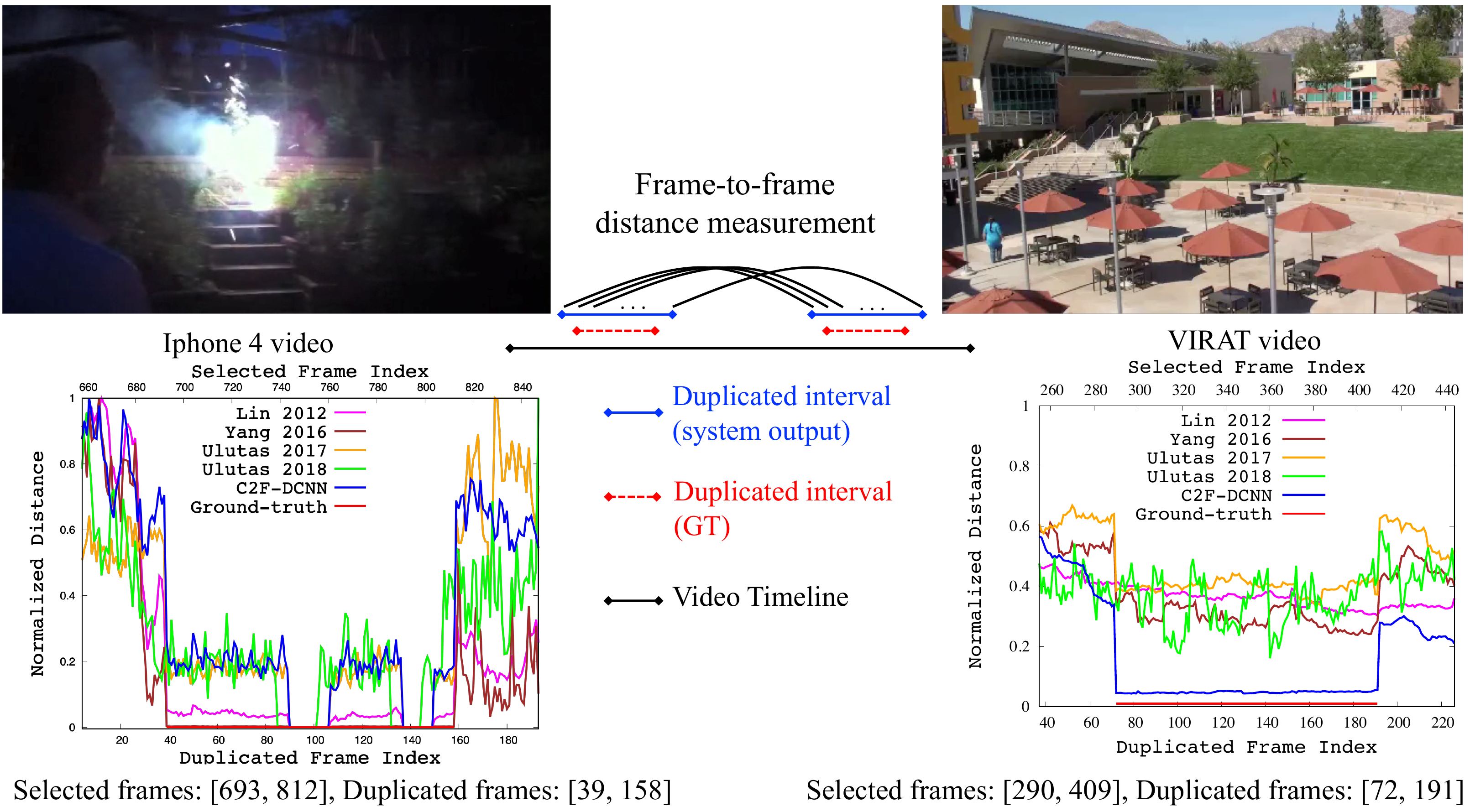}
\end{center}
\vspace{-0.5cm}
   \caption{ Illustration of frame-to-frame distance between duplicated frames and the selected frames.}
\label{fig:framedistances}
\vspace{-0.3cm}
\end{figure*}

The MFC18 dataset 
consists of two sub-sets, Dev dataset and Eval dataset, which we denote as the MFC18-Dev dataset and the MFC18-Eval dataset, respectively. 
There are 231 videos in the MFC18-Dev dataset and 1036 videos in the MFC18-Eval dataset. 
The duration of each video is in the range from 2s to 3 minutes. The frame rate for most of the videos is 29-30 fps, while a smaller number of videos are 10 or 60 fps and only 5 videos are with larger than 240 fps. We opt out 2 videos which have less than 17 frames because the input for the I3D network should have at least 17 frames. We also opt out those 5 videos with large frame rates ($>$220 FPS), since the frame rate for our training videos is not so high. We use the remaining 1029 videos to conduct the video-level performance evaluation.

The detection task is to detect whether or not a video has been manipulated with frame duplication manipulation, while the localization task to localize the duplicated frames index. For the measurement metrics, we use the performance measures of area under the ROC-curve (AUC) for the detection task, and use the Matthews Correlation Coefficient
\begin{equation}
\text{MCC}=\frac{\text{TP} \times \text{TN}-\text{FP} \times \text{FN}}{\sqrt{(\text{TP}+\text{FP})(\text{TP}+\text{FN})(\text{TN}+\text{FP})(\text{TN}+\text{FN})}}\nonumber
\end{equation}
for localization evaluation, where TP, FP, TN, FN refer to frames which represent true positive, false positive, true negative and false negative, respectively. See~\cite{guan2019mfc} for further details on the metrics.

\subsection{Frame-level performance on our self-collected dataset}
To better verify the effectiveness of deep learning solution in frame-duplication detection on the self-collected dataset, we consider four baselines: Lin {\em et al.}'s method~\cite{lin2012detection} that uses histogram difference as the detection features, Yang {\em et al.}'s method~\cite{yang2016using} that is an effective similarity-analysis-based method with SVD features, Ulutas {\em et al.}'s method~\cite{ulutas2017frame} based on binary features and another method by them~\cite{ulutas2018frame} that uses bag-of-words with 130-dimensional SIFT descriptors. Unlike our proposed C2F-DCNN method, all of these methods use traditional feature extraction without deep learning.

Note that the manipulated videos are generated by us, hence both selected original frames and duplicated frames are accessible to us. We treat these experiments as a white-box attack and evaluate the performance of frame-to-frame distance measurements. 



\begin{table}[h]
\caption{The AUC performance of frame-to-frame distance measurements for frame duplication detection on our self-collected video dataset.(unit: \%)}
\vspace{-0.5cm}
\begin{center}
\begin{tabular}{|c|c|c|c|c|}
\hline
Method & Iphone 4 videos & VIRAT videos \\
\hline
Lin 2012~\cite{lin2012detection} & 80.81 & 80.75 \\
Yang 2016~\cite{yang2016using} & 73.79 &  82.13\\
Ulutas 2017~\cite{ulutas2017frame} & 70.46 & 81.32 \\
Ulutas 2018~\cite{ulutas2018frame} & 73.25 & 69.10 \\
\hline
C2F-DCNN  & {\bf 81.46} & {\bf 84.05} \\
\hline
\end{tabular}
\end{center}
\label{tab:auc_x264}
\vspace{-0.5cm}
\end{table}

We run the proposed C2F-DCNN approach and the above mentioned four state-of-the-art approaches on our self-collected dataset 
and the results are summarized in 
Table~\ref{tab:auc_x264}. As we can see,
due to the X264 codec, the contents of the duplicated frames have
been affected so that the detection of a duplicated frame and its corresponding selected frame is very challenging. 
In this case, our C2F-DCNN method still outperforms the four preivous methods. 


To help the reader better understand the comparison, we provide a visualization of the normalized distances between the selected frames and the duplicated frames in Figure~\ref{fig:framedistances}. We can see our C2F-DCNN performs the best for both sample videos, especially with respect to the ability to distinguish the temporal boundary between duplicated frames and non-duplicated frames. All these observations strongly demonstrate the effectiveness of this deep learning approach for frame duplication detection.

\subsection{Video-level performance on the MFC18 dataset}
It is worth mentioning that the duplicated videos in the MFC18 dataset usually include multiple manipulations, and this makes the content between the selected original frames and duplicated frames are affected more or less. Therefore, the testing video in both the MFC18-Dev and the MFC18-Eval datasets are very challenging. Since we are not aware of the details about the generation of all the testing videos, we take this dataset as a black-box attack and evaluate its video-level detection and localization performance.

\begin{figure}
\begin{center}
\includegraphics[width=0.485\textwidth]{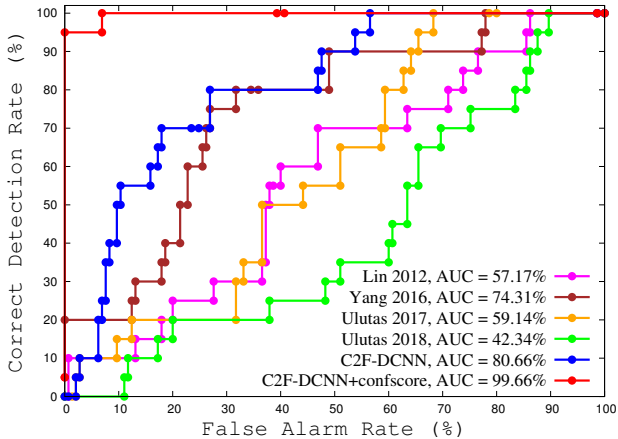}
\end{center}
\vspace{-0.5cm}
   \caption{The ROC curves 
   for video-level frame duplication detection  on the MFC18-Dev dataset.}
\label{fig:det_mfc18_dev}
\vspace{-0.3cm}
\end{figure}
\begin{figure}
\begin{center}
\includegraphics[width=0.485\textwidth]{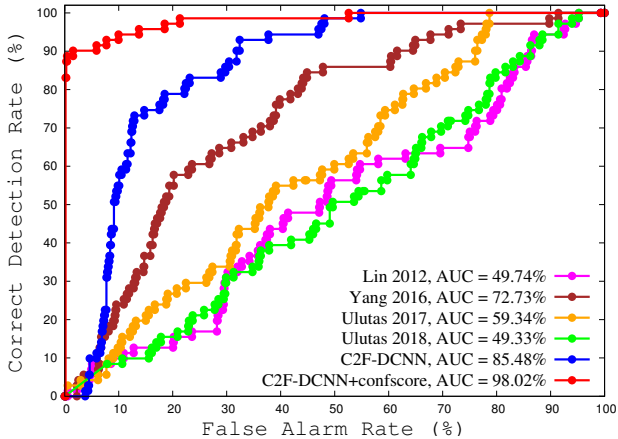}
\end{center}
\vspace{-0.5cm}
   \caption{The ROC curves 
   for video-level frame duplication detection on the MFC18-Eval dataset.}
\label{fig:det_mfc18_eval}
\vspace{-0.3cm}
\end{figure}

We compare the proposed C2F-DCNN method and the above-mentioned four state-of-the-art methods, {\em i.e.}, ``Lin 2012", ``Yang 2016", ``Ulutas 2017" and ``Ulutas 2018" on these two datasets. We use the negative minimum distance ({\em i.e.}, $-d_{min}$) as a default video-level scoring method to generate a video-level score for each competing method, including ours. ``C2F-DCNN+confscore" denotes our best configuration with C2F-DCNN along with the proposed video-level confidence score defined in Equation~\ref{eqn:vconfscore}. In contrast, ``C2F-DCNNa'' uses only $-d_{min}$ as the confidence score. The comparative manipulated video detection results are summarized in Figures~\ref{fig:det_mfc18_dev} and \ref{fig:det_mfc18_eval}. 



A few observations that we would like to point out: (1) C2F-DCNN always outperforms the four previous methods for the video-level frame duplication, with the video-level score as negative minimum distance;
(2) with ``+conf score", our ``C2F-DCNN+confscore" method generates a significant boost in AUC as compared to the baseline score of $-d_min$ and achieves a high correct detection rate at low false alarm rate;
and (3) the proposed ``C2F-DCNN+confscore" method achieves very high AUC scores on the two benchmark datasets: 99.66\% on MFC18-Dev, and 98.02\% on MFC18-Eval.

\begin{table}[h]
\caption{The MCC metric in [-1.0, 1.0] range for video temporal localization on the MFC18 dataset. Our approach generates the best MCC score, where 1.0 is perfect.}
\begin{center}
\vspace{-0.5cm}
\begin{tabular}{|c|c|c|c|c|}
\hline
Method & MFC18-Dev & MFC18-Eval \\
\hline
Lin 2012~\cite{lin2012detection} & 0.2277 & 0.1681 \\
Yang 2016~\cite{yang2016using} & 0.1449 & 0.1548 \\
Ulutas 2017~\cite{ulutas2017frame} & 0.2810 & 0.3147 \\
Ulutas 2018~\cite{ulutas2018frame} & 0.0115 & 0.0391 \\
\hline
C2F-DCNN w/ ResNet  & 0.4618 & 0.3234 \\
C2F-DCNN w/ C3D  & 0.6028 & 0.3488 \\
C2F-DCNN w/ I3D  & {\bf 0.6612} & {\bf 0.3606} \\
\hline
\end{tabular}
\vspace{-0.75cm}
\end{center}
\label{tab:mcc_mfc2018}
\end{table}

\begin{table}[h]
\caption{The video temporal localization performance on the MFC18 dataset. Note $\surd$, $\times$ and $\otimes$ indicate correct cases, incorrect cases and ambiguously incorrect cases, respectively. And $\#(.)$ indicates the number of a kind of specific cases.}
\begin{center}
\vspace{-0.5cm}
\begin{tabular}{|c|c|c|c|}
\hline
Dataset & $\#(\surd)$ & $\#(\times)$ & $\#(\otimes)$ \\
\hline
MFC18-Dev & 14 & 6 & 1 \\
\hline
MFC18-Eval & 33 & 38 & 15 \\
\hline
\end{tabular}
\vspace{-0.5cm}
\end{center}
\label{tab:statics_mfc18}
\end{table}

\begin{figure*}[ht]
\vspace{-0.10in}
\begin{minipage}{0.5\textwidth}
\begin{center}
\subfloat[Completely correct case (0 frame missed).]{\label{fig:complete_correct}
\includegraphics[height=0.09\textheight, width=1.1\textwidth, angle=0]{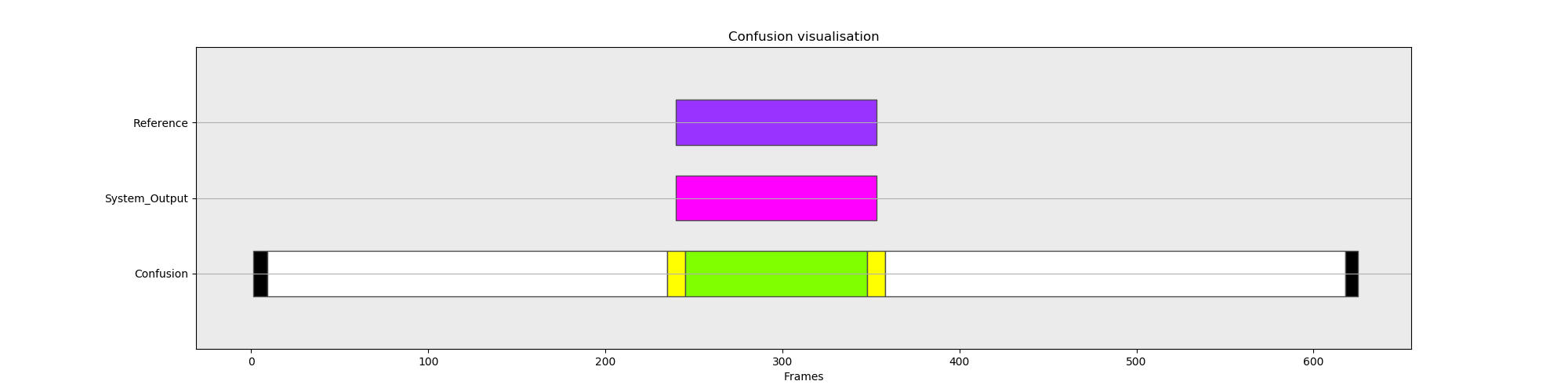}}\\
\end{center}
\end{minipage}
\begin{minipage}{0.5\textwidth}
\begin{center}
\subfloat[Partially correct case  (4 frames missed on the right end only).]{\label{fig:incomplete_correct_a}
\includegraphics[height=0.09\textheight, width=1.1\textwidth, angle=0]{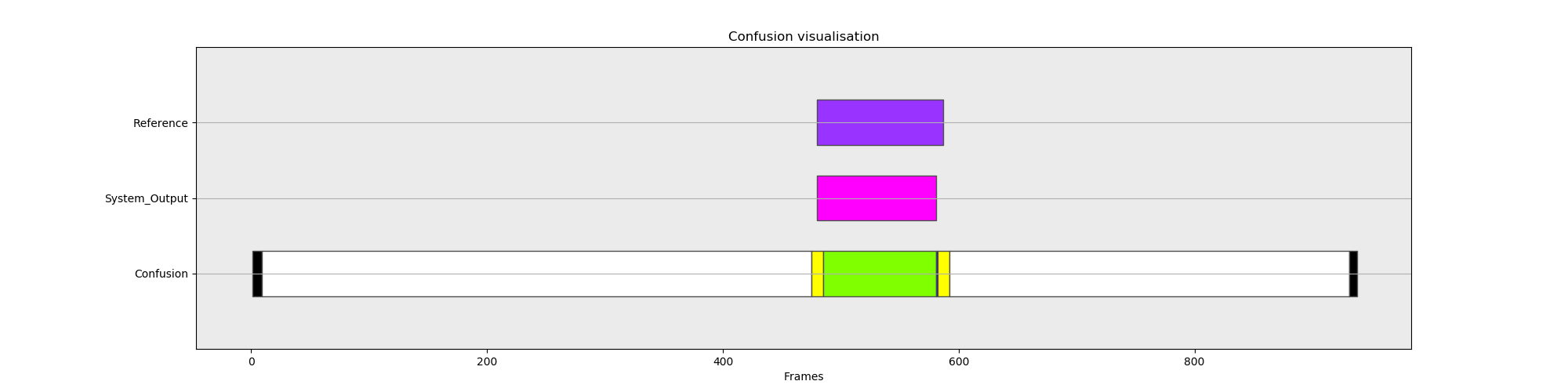}}
\end{center}
\end{minipage}\\
\begin{minipage}{0.5\textwidth}
\begin{center}
\subfloat[Partially correct case (4 frames missed on the left end only).]{\label{fig:incomplete_correct_b}
\includegraphics[height=0.09\textheight, width=1.1\textwidth, angle=0]{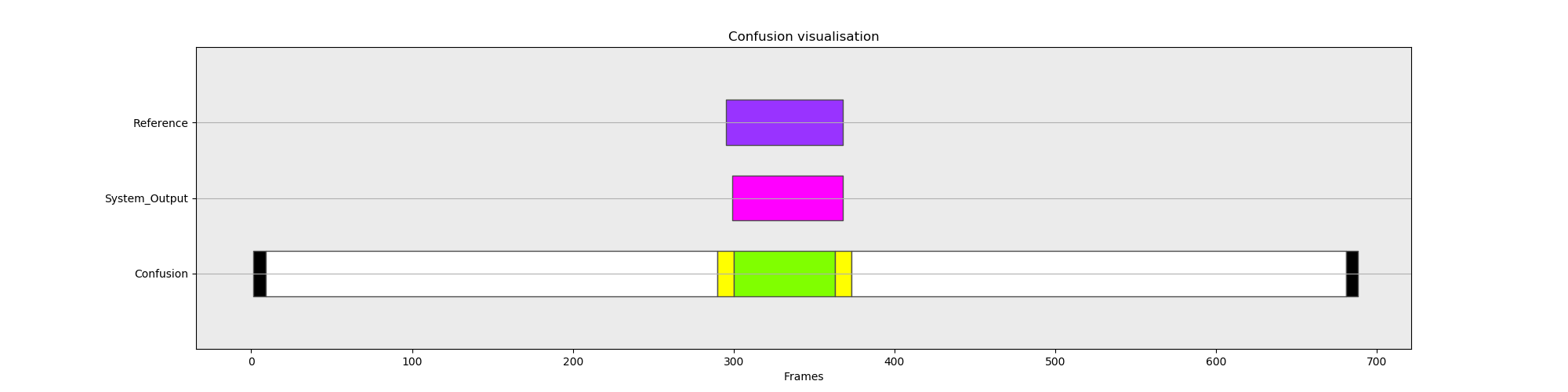}}
\end{center}
\end{minipage}
\begin{minipage}{0.5\textwidth}
\begin{center}
\subfloat[Partially correct case (7 and 4 frames missed on the left and right end).]{\label{fig:incomplete_correct_c}
\includegraphics[height=0.09\textheight, width=1.1\textwidth, angle=0]{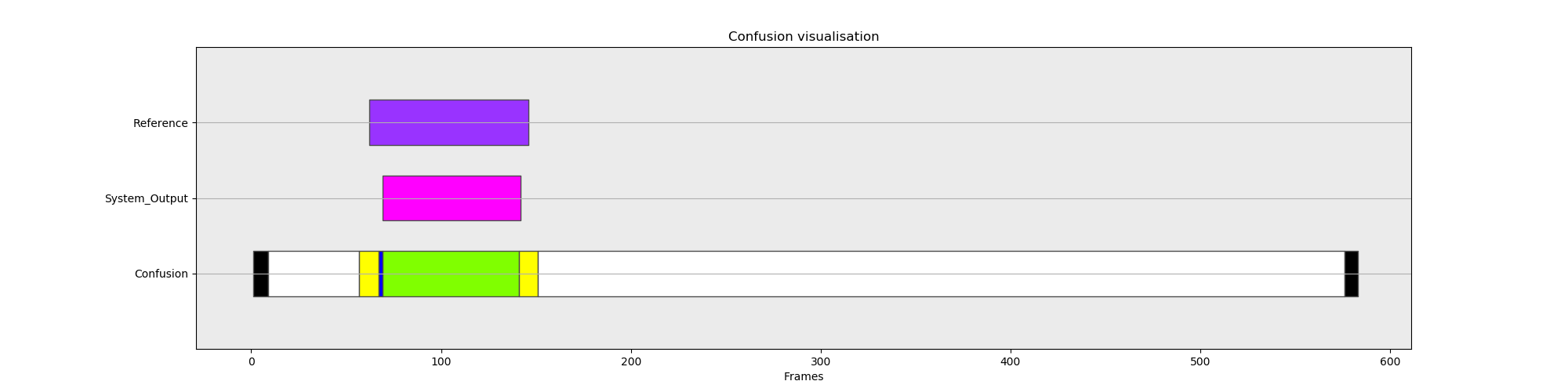}}
\end{center}
\end{minipage}\\
\begin{minipage}{0.5\textwidth}
\begin{center}
\subfloat[Incorrect cases (2 frames gap).]{\label{fig:complete_incorrect}
\includegraphics[height=0.09\textheight, width=1.1\textwidth, angle=0]{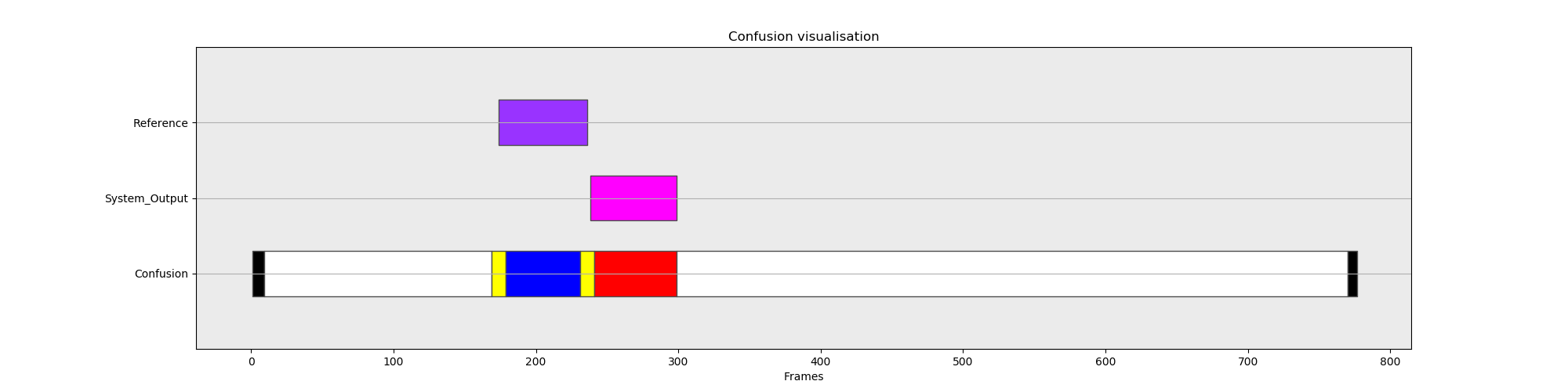}}
\end{center}
\end{minipage}
\begin{minipage}{0.5\textwidth}
\begin{center}
\subfloat[Abmiguously Incorrect case (0 frame gap). ]{\label{fig:ambiguous_incorrect}
\includegraphics[height=0.09\textheight, width=1.1\textwidth, angle=0]{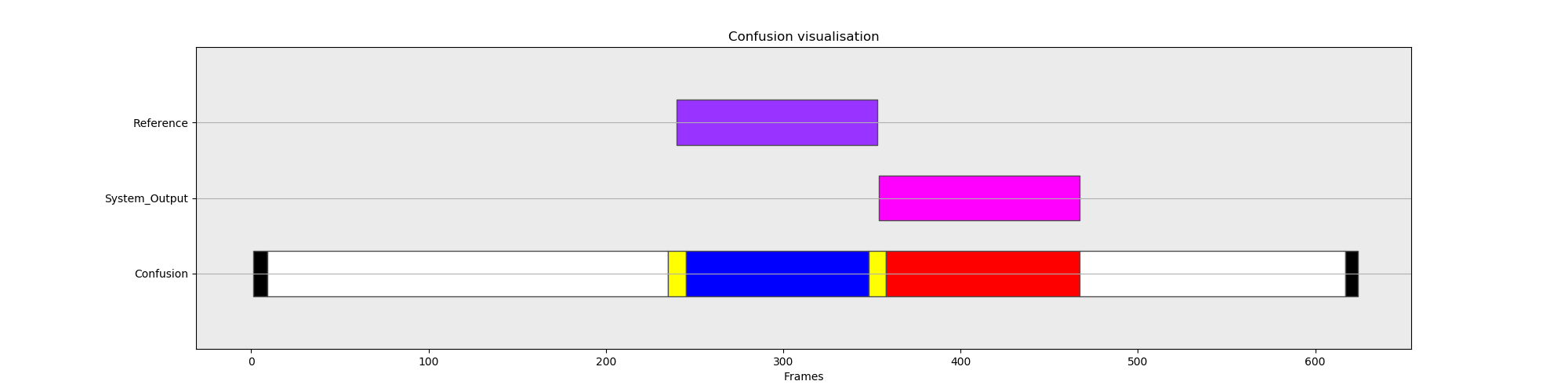}}
\end{center}
\end{minipage}
\vspace{-0.3cm}
\caption{The visualization of confusion bars in video temporal localization. For each subfigure, the top (purple) bar is ground truth indicating duplication, the middle bar (pink) is the system output from the proposed method, and the bottom bar is the confusion calculated based on the above the truth and the system output. Note TN, FN, FP, TP and ``OptOut" in the confusion are marked in white, blue, red, green and yellow / black, respectively. (a) and (b-d) are correct results, which includes completely correct cases and partially correct cases. (e) and (f) show the failure cases.}
\label{fig:loc_mfc18_visualization}
\vspace{-0.3cm}
\end{figure*}

We also performed quantified analysis of the temporal localization within a manipulated video with frame duplication. For comparison with the four previous methods, we use the feature distance between any two consecutive frames. For the proposed C2F-DCNN approach, the best configuration ``C2F-DCNN w/ I3D'' includes the I3D network as the inconsistency detector. We also provide two baseline variants by replacing the I3D inconsistency detector with a ResNet network feature distance $S_{Res}(i)$ only (``C2F-DCNN w/ ResNet'') or the C3D network's scores $ S_{C3D}^{drop}(i) - \lambda S_{C3D}^{none}(i)$ from~\cite{Long:CVPRW2017} (``C2F-DCNN w/ C3D''). 
The temporal localization results are summarized in Table~\ref{tab:mcc_mfc2018}, from which we can observe that (1) our deep learning solutions, ``C2F-DCNN w/ ResNet", ``C2F-DCNN w/ C3D" or ``C2F-DCNN w/ I3D" work better than the four previous methods and ``C2F-DCNN w/ I3D" performs the best. 
These observations suggest that 3D convolutional kernel is able to measure the inconsistency between the consecutive frames, and both RGB data stream and optical flow data stream are complementary to further improve the performance.  

To better understand the video temporal localization measurement, we plot the confusion bars on the video timeline based on the truth and the corresponding system output under different scenarios, as shown in Figure~\ref{fig:loc_mfc18_visualization}. We would like to emphasize that no algorithm is able to distinguish duplicated frames from selected frames for the ambiguously incorrect cases indicated as $\otimes$ in Table~\ref{tab:statics_mfc18}, because such videos often break the assumption of temporal consistency and in many cases the duplicated frames are difficult to identify by naked eye. 


\subsection{Discussion}
Multiple factors cause frame duplication detection and localization becoming more and more challenging in video forgries. These factors includes high frame rates, multiple manipulations ({\em e.g.}, ``SelectCutFrames", ``TimeAlterationWarp", ``AntiForensicCopyExif", ``RemoveCamFingerprintPRNU"~\footnote{These manipulation operation are defined in the MFC18 dataset.}) involved before and after, and gaps between the selected frames and the duplicated frames. In particular, zero gap between the selected frames and the duplicated frames render the manipulation undetectable because the inconsistency which should exist in end of the duplicated frames does not appear in the video temporal context.


Regarding the runtime, the I3D network for inconsistency detection is the most expensive component in our framework but we only apply it on the candidate frames that are likely to have frame duplication manipulations detected in the coarse-search stage. 
For each testing video clip with a 16-frame length, it takes about 2 seconds with our learned I3D network. For a one-minute short video with 30 FPS, it requires less than 5 minutes to complete the testing throughout all the frame sequence.

\section{Conclusions and Future Work}
In this paper, we propose a coarse-to-fine deep learning approach for frame duplication detection at both frame-level and video-level, as well as for the video temporal localization. We also propose a heuristic strategy to formulate the video-level confidence score, as well as an I3D network based inconsistency detector to distinguish the duplicated frames from the selected frames. The experimental results have demonstrate the robustness and effectiveness of the proposed method.

Our future work includes continuing to extend multi-stream 3D neural networks for both frame drop, frame duplication and other video manipulation tasks like looping detection, working on frame-rate variations, and train on multiple manipulations, investigating the effects of various video codecs on algorithm accuracy. 


{\small
\bibliographystyle{ieee}
\bibliography{CopyPaste}

\begin{thebibliography}{10}\itemsep=-1pt

\bibitem{Aneja_2018_CVPR}
J.~Aneja, A.~Deshpande, and A.~G. Schwing.
\newblock Convolutional image captioning.
\newblock In {\em The IEEE Conference on Computer Vision and Pattern
  Recognition (CVPR)}, June 2018.

\bibitem{carreira2017quo}
J.~Carreira and A.~Zisserman.
\newblock Quo vadis, action recognition? a new model and the kinetics dataset.
\newblock In {\em Computer Vision and Pattern Recognition (CVPR), 2017 IEEE
  Conference on}, pages 4724--4733. IEEE, 2017.

\bibitem{Chen_2018_CVPR}
Y.~Chen, W.~Li, C.~Sakaridis, D.~Dai, and L.~Van~Gool.
\newblock Domain adaptive faster r-cnn for object detection in the wild.
\newblock In {\em The IEEE Conference on Computer Vision and Pattern
  Recognition (CVPR)}, June 2018.

\bibitem{christlein2012evaluation}
V.~Christlein, C.~Riess, J.~Jordan, C.~Riess, and E.~Angelopoulou.
\newblock An evaluation of popular copy-move forgery detection approaches.
\newblock {\em arXiv preprint arXiv:1208.3665}, 2012.

\bibitem{DouzeGJMS08}
M.~Douze, A.~Gaidon, H.~Jegou, M.~Marszalek, and C.~Schmid.
\newblock Inria-lear's video copy detection system.
\newblock In {\em {TRECVID} 2008 workshop participants notebook papers,
  Gaithersburg, MD, USA, November 2008}, 2008.

\bibitem{d2019patchmatch}
L.~D’Amiano, D.~Cozzolino, G.~Poggi, and L.~Verdoliva.
\newblock A patchmatch-based dense-field algorithm for video copy--move
  detection and localization.
\newblock {\em IEEE Transactions on Circuits and Systems for Video Technology},
  29(3):669--682, 2019.

\bibitem{guan2019mfc}
H.~Guan, M.~Kozak, E.~Robertson, Y.~Lee, A.~N. Yates, A.~Delgado, D.~Zhou,
  T.~Kheyrkhah, J.~Smith, and J.~Fiscus.
\newblock Mfc datasets: Large-scale benchmark datasets for media forensic
  challenge evaluation.
\newblock In {\em 2019 IEEE Winter Applications of Computer Vision Workshops
  (WACVW)}, pages 63--72. IEEE, 2019.

\bibitem{Hara_2018_CVPR}
K.~Hara, H.~Kataoka, and Y.~Satoh.
\newblock Can spatiotemporal 3d cnns retrace the history of 2d cnns and
  imagenet?
\newblock In {\em The IEEE Conference on Computer Vision and Pattern
  Recognition (CVPR)}, June 2018.

\bibitem{He_2016_CVPR}
K.~He, X.~Zhang, S.~Ren, and J.~Sun.
\newblock Deep residual learning for image recognition.
\newblock In {\em The IEEE Conference on Computer Vision and Pattern
  Recognition (CVPR)}, June 2016.

\bibitem{hu2012improved}
Y.~Hu, C.-T. Li, Y.~Wang, and B.-b. Liu.
\newblock An improved fingerprinting algorithm for detection of video frame
  duplication forgery.
\newblock {\em International Journal of Digital Crime and Forensics (IJDCF)},
  4(3):20--32, 2012.

\bibitem{huang2018multi}
T.~Huang, X.~Zhang, W.~Huang, L.~Lin, and W.~Su.
\newblock A multi-channel approach through fusion of audio for detecting video
  inter-frame forgery.
\newblock {\em Computers \& Security}, 77:412--426, 2018.

\bibitem{karpathy2014large}
A.~Karpathy, G.~Toderici, S.~Shetty, T.~Leung, R.~Sukthankar, and L.~Fei-Fei.
\newblock Large-scale video classification with convolutional neural networks.
\newblock In {\em Proceedings of the IEEE conference on Computer Vision and
  Pattern Recognition}, pages 1725--1732, 2014.

\bibitem{KawaiSY07}
Y.~Kawai, H.~Sumiyoshi, and N.~Yagi.
\newblock Shot boundary detection at {TRECVID} 2007.
\newblock In {\em {TRECVID} 2007 workshop participants notebook papers,
  Gaithersburg, MD, USA, November 2007}, 2007.

\bibitem{law2006robust}
J.~Law-To, O.~Buisson, V.~Gouet-Brunet, and N.~Boujemaa.
\newblock Robust voting algorithm based on labels of behavior for video copy
  detection.
\newblock In {\em Proceedings of the 14th ACM international conference on
  Multimedia}, pages 835--844. ACM, 2006.

\bibitem{le2010national}
D.-D. Le, S.~Poullot, X.~Wu, B.~Nouvel, and S.~Satoh.
\newblock National institute of informatics, japan at trecvid 2010.
\newblock In {\em TRECVID}, 2010.

\bibitem{lin2012detection}
G.-S. Lin and J.-F. Chang.
\newblock Detection of frame duplication forgery in videos based on spatial and
  temporal analysis.
\newblock {\em International Journal of Pattern Recognition and Artificial
  Intelligence}, 26(07):1250017, 2012.

\bibitem{Long:CVPRW2017}
C.~Long, E.~Smith, A.~Basharat, and A.~Hoogs.
\newblock A c3d-based convolutional neural network for frame dropping detection
  in a single video shot.
\newblock In {\em {IEEE} International Conference on Computer Vision and
  Pattern Recognition Workshop (CVPR-W) on Media Forensics}, 2017.

\bibitem{milani2012overview}
S.~Milani, M.~Fontani, P.~Bestagini, M.~Barni, A.~Piva, M.~Tagliasacchi, and
  S.~Tubaro.
\newblock An overview on video forensics.
\newblock {\em APSIPA Transactions on Signal and Information Processing}, 1,
  2012.

\bibitem{mohamadian2013detection}
Z.~Mohamadian and A.~A. Pouyan.
\newblock Detection of duplication forgery in digital images in uniform and
  non-uniform regions.
\newblock In {\em Computer Modelling and Simulation (UKSim), 2013 UKSim 15th
  International Conference on}, pages 455--460. IEEE, 2013.

\bibitem{oh2011large}
S.~Oh, A.~Hoogs, A.~Perera, N.~Cuntoor, C.-C. Chen, J.~T. Lee, S.~Mukherjee,
  J.~Aggarwal, H.~Lee, L.~Davis, et~al.
\newblock A large-scale benchmark dataset for event recognition in surveillance
  video.
\newblock In {\em Computer vision and pattern recognition (CVPR), 2011 IEEE
  conference on}, pages 3153--3160. IEEE, 2011.

\bibitem{NIPS2015_5638}
S.~Ren, K.~He, R.~Girshick, and J.~Sun.
\newblock Faster r-cnn: Towards real-time object detection with region proposal
  networks.
\newblock In C.~Cortes, N.~D. Lawrence, D.~D. Lee, M.~Sugiyama, and R.~Garnett,
  editors, {\em Advances in Neural Information Processing Systems (NIPS)},
  pages 91--99. 2015.

\bibitem{sowmya2015survey}
K.~Sowmya and H.~Chennamma.
\newblock A survey on video forgery detection.
\newblock {\em International Journal of Computer Engineering and Applications},
  9(2):17--27, 2015.

\bibitem{Stock_2018_ECCV}
P.~Stock and M.~Cisse.
\newblock Convnets and imagenet beyond accuracy: Understanding mistakes and
  uncovering biases.
\newblock In {\em The European Conference on Computer Vision (ECCV)}, September
  2018.

\bibitem{Tang_2018_ECCV}
P.~Tang, X.~Wang, A.~Wang, Y.~Yan, W.~Liu, J.~Huang, and A.~Yuille.
\newblock Weakly supervised region proposal network and object detection.
\newblock In {\em The European Conference on Computer Vision (ECCV)}, September
  2018.

\bibitem{tran2015learning}
D.~Tran, L.~Bourdev, R.~Fergus, L.~Torresani, and M.~Paluri.
\newblock Learning spatiotemporal features with 3d convolutional networks.
\newblock In {\em Proceedings of the IEEE international conference on computer
  vision}, pages 4489--4497, 2015.

\bibitem{ulutas2017frame}
G.~Ulutas, B.~Ustubioglu, M.~Ulutas, and V.~Nabiyev.
\newblock Frame duplication/mirroring detection method with binary features.
\newblock {\em IET Image Processing}, 11(5):333--342, 2017.

\bibitem{ulutas2018frame}
G.~Ulutas, B.~Ustubioglu, M.~Ulutas, and V.~V. Nabiyev.
\newblock Frame duplication detection based on bow model.
\newblock {\em Multimedia Systems}, pages 1--19, 2018.

\bibitem{VenugopalanXDRMS14}
S.~Venugopalan, H.~Xu, J.~Donahue, M.~Rohrbach, R.~J. Mooney, and K.~Saenko.
\newblock Translating videos to natural language using deep recurrent neural
  networks.
\newblock In {\em North American Chapter of the Association for Computational
  Linguistics – Human Language Technologies (NAACL-HLT)}, 2015.

\bibitem{wang2009fast}
J.~Wang, G.~Liu, Z.~Zhang, Y.~Dai, and Z.~Wang.
\newblock Fast and robust forensics for image region-duplication forgery.
\newblock {\em Acta Automatica Sinica}, 35(12):1488--1495, 2009.

\bibitem{wang2014videoJCC}
Q.~Wang, Z.~Li, Z.~Zhang, and Q.~Ma.
\newblock Video inter-frame forgery identification based on consistency of
  correlation coefficients of gray values.
\newblock {\em Journal of Computer and Communications}, 2(04):51, 2014.

\bibitem{wang2014videoST}
Q.~Wang, Z.~Li, Z.~Zhang, and Q.~Ma.
\newblock Video inter-frame forgery identification based on optical flow
  consistency.
\newblock {\em Sensors \& Transducers}, 166(3):229, 2014.

\bibitem{wang2007exposing}
W.~Wang and H.~Farid.
\newblock Exposing digital forgeries in video by detecting duplication.
\newblock In {\em Proceedings of the 9th workshop on Multimedia \& security},
  pages 35--42. ACM, 2007.

\bibitem{wang2018robust}
X.-y. Wang, Y.-n. Liu, H.~Xu, P.~Wang, and H.-y. Yang.
\newblock Robust copy--move forgery detection using quaternion exponent
  moments.
\newblock {\em Pattern Analysis and Applications}, 21(2):451--467, 2018.

\bibitem{wu2014exposing}
Y.~Wu, X.~Jiang, T.~Sun, and W.~Wang.
\newblock Exposing video inter-frame forgery based on velocity field
  consistency.
\newblock In {\em Acoustics, speech and signal processing (ICASSP), 2014 IEEE
  International Conference on}, pages 2674--2678. IEEE, 2014.

\bibitem{Xie_2018_ECCV}
S.~Xie, C.~Sun, J.~Huang, Z.~Tu, and K.~Murphy.
\newblock Rethinking spatiotemporal feature learning: Speed-accuracy trade-offs
  in video classification.
\newblock In {\em The European Conference on Computer Vision (ECCV)}, September
  2018.

\bibitem{yang2018copy}
B.~Yang, X.~Sun, H.~Guo, Z.~Xia, and X.~Chen.
\newblock A copy-move forgery detection method based on cmfd-sift.
\newblock {\em Multimedia Tools and Applications}, 77(1):837--855, 2018.

\bibitem{yang2016using}
J.~Yang, T.~Huang, and L.~Su.
\newblock Using similarity analysis to detect frame duplication forgery in
  videos.
\newblock {\em Multimedia Tools and Applications}, 75(4):1793--1811, 2016.

\bibitem{Yu_2018_CVPR}
H.~Yu, S.~Cheng, B.~Ni, M.~Wang, J.~Zhang, and X.~Yang.
\newblock Fine-grained video captioning for sports narrative.
\newblock In {\em The IEEE Conference on Computer Vision and Pattern
  Recognition (CVPR)}, June 2018.

\bibitem{zhao2018inter}
D.-N. Zhao, R.-K. Wang, and Z.-M. Lu.
\newblock Inter-frame passive-blind forgery detection for video shot based on
  similarity analysis.
\newblock {\em Multimedia Tools and Applications}, pages 1--20, 2018.

\end{thebibliography}
}

\end{document}